\documentclass[a4paper,twoside]{article}

\usepackage{epsfig}
\usepackage{subcaption}
\usepackage{xcolor}
\usepackage{calc}
\usepackage{amssymb}
\usepackage{booktabs}
\usepackage{amstext}
\usepackage{amsmath}
\usepackage{multirow}
\usepackage{amsthm}
\usepackage{multicol}
\usepackage{pslatex}
\usepackage{apalike}
\usepackage{bm}
\usepackage{arydshln}
\usepackage[linesnumbered,ruled,vlined]{algorithm2e}
\usepackage{hyperref}
\usepackage{mathrsfs}
\usepackage[bottom]{footmisc}
\usepackage{SCITEPRESS}

\hypersetup{
    colorlinks=true,
    citecolor=green
}

\begin{document}

\title{A Re-ranking Method using K-nearest Weighted Fusion\\
for Person Re-identification}

\author{\authorname{Quang-Huy Che\sup{1,}\sup{2,}\thanks{First two authors contribute equally}, Le-Chuong Nguyen\sup{1,}\sup{2,}\footnotemark[1], Gia-Nghia Tran\sup{1,}\sup{2},
\\Dinh-Duy Phan\sup{1,}\sup{2} and Vinh-Tiep Nguyen\sup{1,}\sup{2,}\thanks{Corresponding author}}
\affiliation{\sup{1}University of Information Technology, Ho Chi Minh City, Vietnam}
\affiliation{\sup{2}Vietnam National University, Ho Chi Minh City, Vietnam}
\email{huycq@uit.edu.vn, 21520655@gm.uit.edu.vn,\{nghiatg, duypd, tiepnv\}@uit.edu.vn}
}


\abstract{In person re-identification, re-ranking is a crucial step to enhance the overall accuracy by refining the initial ranking of retrieved results. Previous studies have mainly focused on features from single-view images, which can cause view bias and issues like pose variation, viewpoint changes, and occlusions. Using multi-view features to present a person can help reduce view bias. In this work, we present an efficient re-ranking method that generates multi-view features by aggregating neighbors' features using \textit{K-nearest Weighted Fusion} (\textit{KWF}) method. Specifically, we hypothesize that features extracted from re-identification models are highly similar when representing the same identity. Thus, we select \textbf{K} neighboring features in an unsupervised manner to generate multi-view features. Additionally, this study explores the weight selection strategies during feature aggregation, allowing us to identify an effective strategy. Our re-ranking approach does not require model fine-tuning or extra annotations, making it applicable to large-scale datasets. We evaluate our method on the person re-identification datasets Market1501, MSMT17, and Occluded-DukeMTMC. The results show that our method significantly improves Rank@1 and mAP when re-ranking the top \textbf{M} candidates from the initial ranking results. Specifically, compared to the initial results, our re-ranking method achieves improvements of \textbf{9.8\%}/\textbf{22.0\%} in Rank@1 on the challenging datasets: MSMT17 and Occluded-DukeMTMC, respectively. Furthermore, our approach demonstrates substantial enhancements in computational efficiency compared to other re-ranking methods. Code is available at \href{https://github.com/chequanghuy/Enhancing-Person-Re-Identification-via-UFFM-and-AMC}{https://github.com/chequanghuy/Enhancing-Person-Re-Identification-via-UFFM-and-AMC}.
}

\onecolumn \maketitle \normalsize \setcounter{footnote}{0} \vfill

\section{\uppercase{Introduction}}
\label{sec:introduction}

\begin{figure}
    \centering
    \includegraphics[width=0.97\linewidth]{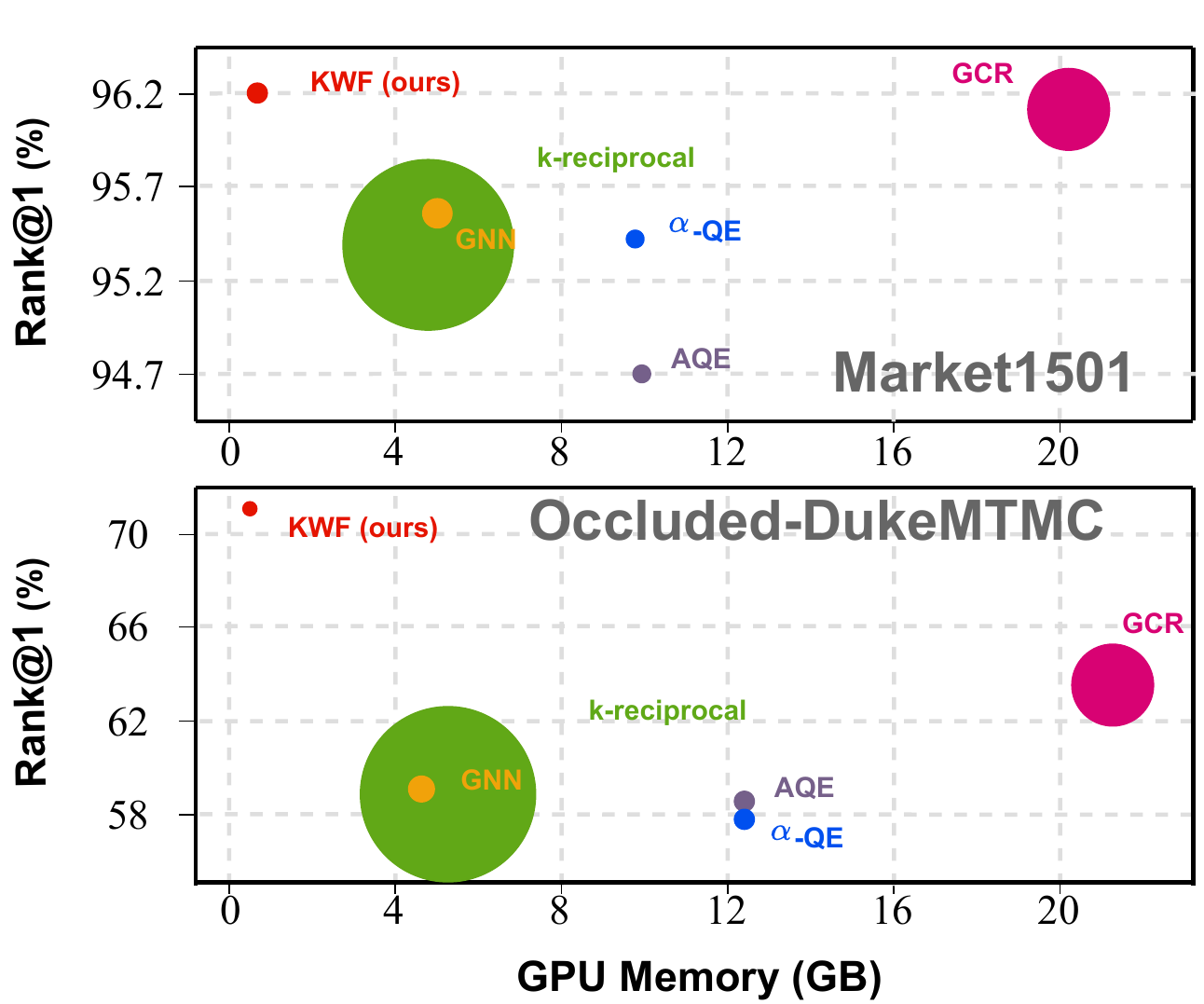}
    \caption{\small{Comparing the computational cost and Rank@1 performance of various reranking methods on the Market1501 and Occluded-DukeMTMC datasets. The y-axis shows Rank@1, the x-axis represents GPU memory usage, and the circle size indicates evaluation time, with larger circles representing longer evaluation times.}}
    \label{fig:compare}
\end{figure}

Person re-identification (ReID) \cite{bot,centroids,part_aware,beyond_app,body_part,clip-reid} is a computer vision task that involves recognizing and matching a person across multiple images or video frames from different cameras. The goal is to reidentify a person despite variations in pose, lighting, camera views, and occlusion. In a typical ReID system, given a query image, the system retrieves and ranks candidate images from a gallery based on their similarity to the query. Initial rankings are based on features from deep learning models and distance metrics. Re-ranking refines this list to improve re-identification accuracy, aiming to give higher ranks to relevant images.

\begin{figure*}
    \centering
    \includegraphics[width=1\linewidth]{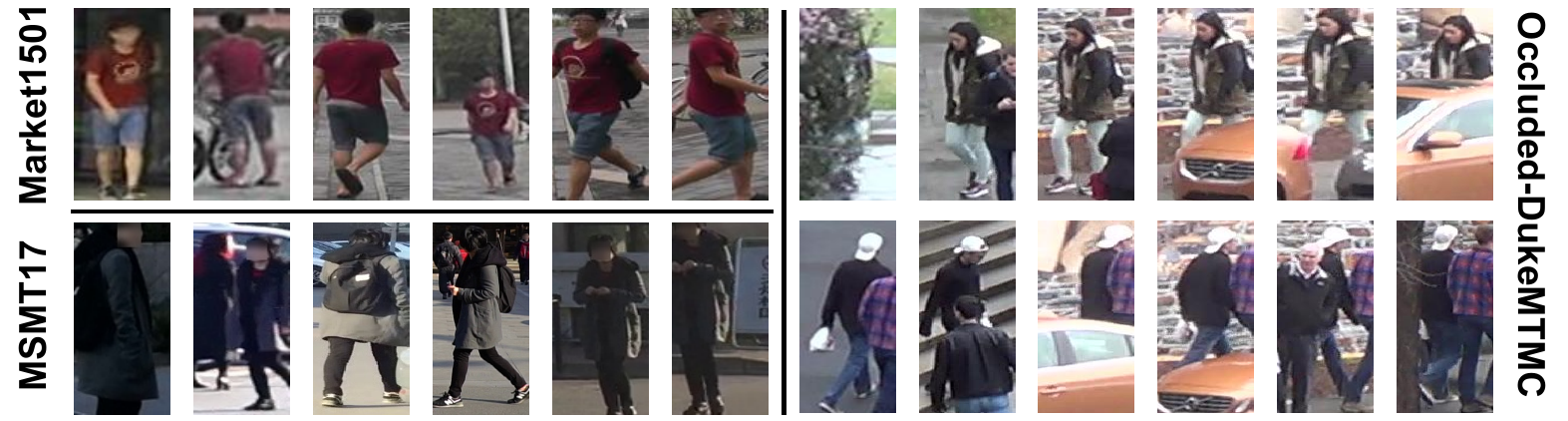}
    \caption{Person re-identification datasets often include eight images of the same person taken from different viewpoints and cameras. The Market1501 and MSMT17 datasets are known for their diverse viewpoints, lighting conditions, and backgrounds.  On the other hand, the Occluded-DukeMTMC dataset poses extra challenges with its complex environments and frequent partial occlusions, complicating the re-identification process.}
    \label{fig:multi_example}
\end{figure*}

Re-ranking is a powerful technique widely employed in various tasks, including re-identification \cite{k_reciprocal,divide_fuse,adaptive_rerank,semi_rerank,multiattn_rerank,GNN_rerank}.
A notable advantage of many re-ranking techniques is their ability to be deployed without needing additional training samples, allowing them to be applied to any initial ranking results. Traditionally, common methods for re-ranking include recalculating image similarities by integrating additional information and using advanced similarity metrics \cite{pairwise_metric,local_maximal,view_adaptive,face_emd}, feature similarity-based methods \cite{alphaQE,aqe}, neighbor similarity \cite{SCE,k_reciprocal,pose_rerank,knn_rerank,hello_neighbor}, and graph-based approaches \cite{GNN_rerank,graph_rerank}. However, previous methods still rely on single-view features directly extracted from the feature extraction model to represent gallery features. Relying only on single-view features can result in a lack of the necessary information to fully and accurately represent pedestrians, especially in complex situations where images are captured from different cameras.

The study \cite{gnn_reid,uffm} emphasized the limitation of view bias in single-view features when re-identifying pedestrians from different camera viewpoints. Since a pedestrian can be captured by multiple cameras, two images of the same person may not have the same details, as one image might lack information present in the other, as illustrated in Figure \ref{fig:multi_example}. Generating features that capture information from multiple viewpoints is an effective solution for mitigating view bias. In this paper, we approach transforming single-view features to multi-view features. We draw several conclusions, including: (1) single-view features can be effectively transformed into multi-view features, (2) selecting single-view features can be done effectively in an unsupervised manner if the correct number of \textbf{K} is chosen, without the need for model fine-tuning, and (3) generating single-view features by selecting appropriate weights allows the generation of multi-view features that better capture the information from single-view features.

In this paper, we propose  generating multi-view features to represent all images in the gallery set during the re-ranking stage, thereby addressing the issue of view bias. To achieve this, we generate multi-view features using the \textit{K-nearest Weighted Fusion (KWF)} method, which generates multi-view features from K-nearest neighbor features. Our proposed unsupervised neighbor feature selection method does not require fine-tuning of pre-trained models. The contributions of this paper are summarized as follows:

\begin{itemize}
    \item We propose two-stage hierarchical person re-identification: the first stage involves ranking based on single-view features, followed by a second stage re-ranking using multi-view features.
    \item We propose the \textit{KWF} method, a multi-view feature representation used during the re-ranking stage. This representation prevents the view bias problem in person re-identification.
    \item We study the effectiveness of weight selection strategies in multi-view feature generation for \textit{KWF}, including \textit{Uniform, Inverse Distance Power} and \textit{Exponential Decay}.
    \item We perform extensive experiments comparing our proposed methods with other re-ranking approaches on datasets like Market-1501, MSMT17, and Occluded-DukeMTMC.
\end{itemize}

\begin{figure*}
    \centering
    \includegraphics[width=1\linewidth]{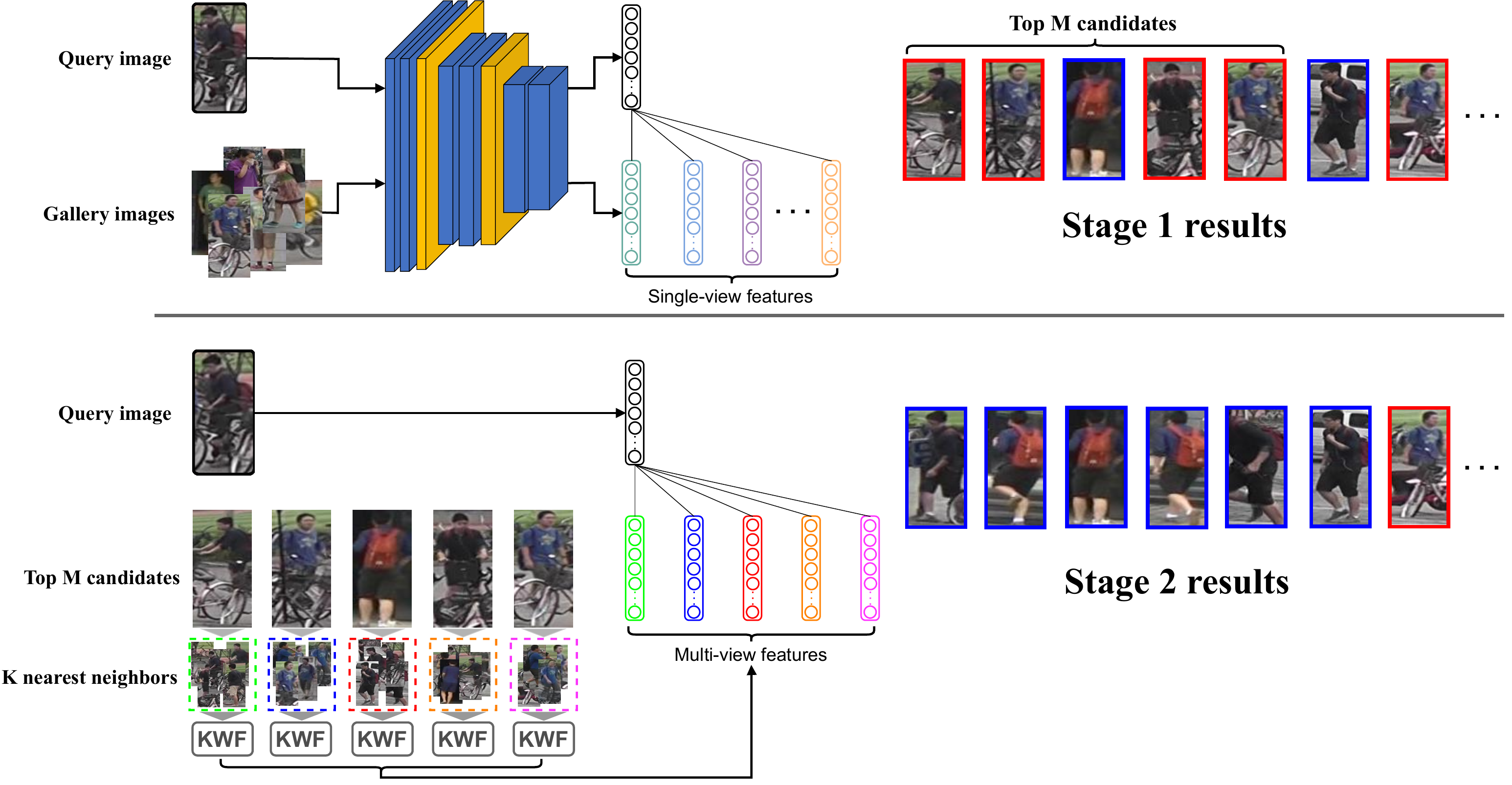}
    \caption{Our person identification pipeline consists of two stages. In the first stage, gallery images are ranked according to the cosine distance between the query and gallery images. In second stage, the top 5 candidates from the first stage are re-ranked using multi-view features.}
    \label{fig:pipeline}
\end{figure*}

\section{\uppercase{Related Work}}

\subsection{Re-ranking approach} Image retrieval in general, and person re-identification in particular, involves searching for images from large databases based on a query image. Re-ranking techniques are crucial in refining the initial search results to enhance retrieval accuracy. Among these, feature similarity-based methods \cite{aqe,alphaQE,DQE} leverage the nearest samples with similar features to enrich the query features by aggregating the features of neighbors. Additionally, neighbor similarity-based methods \cite{SCE,k_reciprocal,hello_neighbor} rely on the number of shared neighbors between images, using k-reciprocal nearest neighbors to efficiently exploit the relationships among images. Furthermore, graph-based re-ranking methods \cite{graph_rerank,GNN_rerank} capture the topological structure of the data and refine the learned features, yielding promising results. While some re-ranking methods do not require extra annotations, others require human interaction or label supervision \cite{human_3,human_2,human_1}. Most re-ranking methods focus on directly exploiting relationships between the initial features. However, the effectiveness of enhancing multi-view features in this context remains underexplored. Therefore, in this study, we propose a method that employs multi-view features during re-ranking without requiring model fine-tuning or extra annotations.

\begin{figure}
    \centering
    \includegraphics[width=1\linewidth]{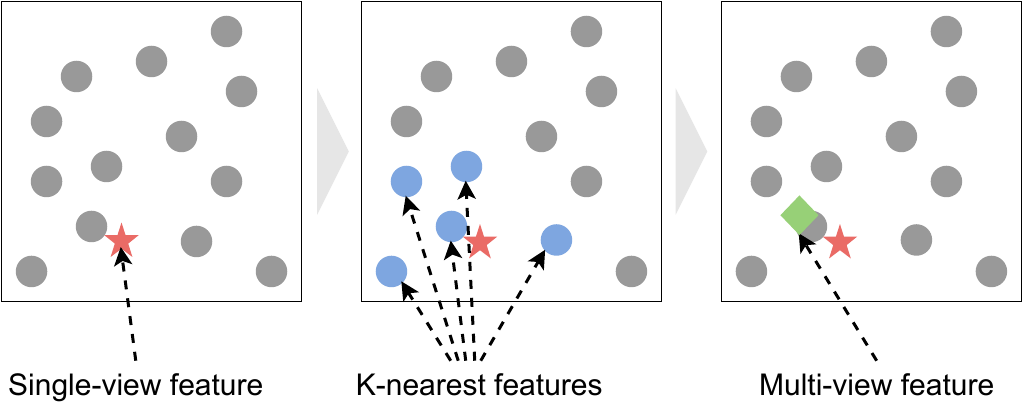}
    \caption{The process of generating multi-view features from single-view features: First, the single-view feature calculates the distance to all features in the gallery to find the \textbf{K} nearest neighbors. Then, the \textit{KWF} method aggregates the neighboring features to generate a multi-view feature.}
    \label{fig:single2multi}
\end{figure}

\subsection{Multi-view feature representation} Previous studies \cite{centroids,gnn_reid} have explored feature fusion in re-identification tasks. In \cite{centroids}, the authors suggested aggregating features of the same class during the query process, which proved effective in both accuracy and query time. However, this approach requires additional labels to select representative features for images of the same class. Therefore, this method requires class-level labeling for images when applied in real-world scenarios, highlighting a limitation of this approach. Yinsong et al. \cite{gnn_reid} proposed feature aggregation using a Graph Neural Network (GNN). While promising, GNN-based methods have drawbacks, such as hardware-dependent computation costs and significant computational and storage demands for large graphs. This study introduces a weighted aggregation method for single-view features based on neighboring features. Our method selects neighboring features unsupervisedly, ensuring efficiency by leveraging the feature extraction capabilities of pre-trained re-identification models.

\begin{figure*}[ht]
    \centering
    
    \begin{subfigure}[b]{0.24\textwidth}
        \centering
        \includegraphics[width=\textwidth]{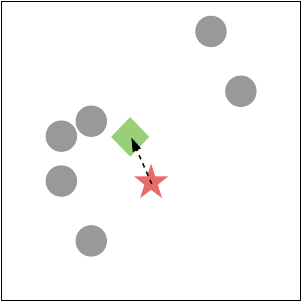} 
        \caption{\scriptsize Uniform}
        \label{fig:sub2}
    \end{subfigure}
    \begin{subfigure}[b]{0.24\textwidth}
        \centering
        \includegraphics[width=\textwidth]{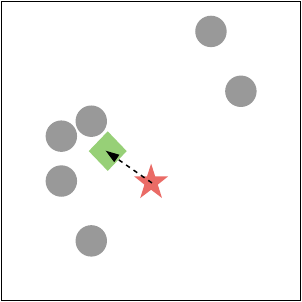} 
        \caption{\scriptsize Inverse Distance Power ($p$ = 1)}
        \label{fig:sub3}
    \end{subfigure}
    \begin{subfigure}[b]{0.24\textwidth}
        \centering
        \includegraphics[width=\textwidth]{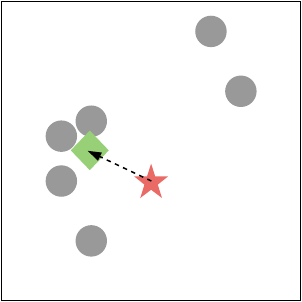} 
        \caption{\scriptsize Inverse Distance Power ($p$ = 2)}
        \label{fig:sub4}
    \end{subfigure}
    \begin{subfigure}[b]{0.24\textwidth}
        \centering
        \includegraphics[width=\textwidth]{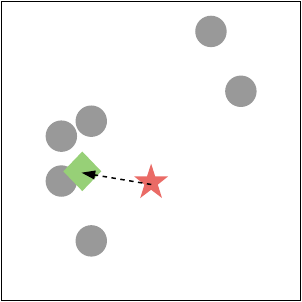} 
        \caption{\scriptsize Exponential Decay}
        \label{fig:sub5}
    \end{subfigure}
    
    \caption{The process involves transforming single-view features into multi-view features using the \textit{KWF} method. In each figure, the \textcolor{red}{red star} represents the single-view feature, \textcolor{gray}{gray circles} represent the nearest neighboring features (with \textbf{K} = 6), and the green quadrilateral represents the generated multi-view feature.}
    \label{fig:weighted}
    
\end{figure*}

\section{\uppercase{Method}}

In this section, we propose a two-stage hierarchical approach for the person re-identification task. In the first stage, images in the gallery are ranked based on their pairwise distance to a given query. To compute this pairwise distance, we use the cosine distance of features extracted by a re-identification model, where the features representing the gallery images are referred to as single-view features. In the second stage, the single-view features extracted from the first stage are transformed using the \textit{KWF} method. This method aggregates the neighboring features of the single-view features to generate multi-view features. To effectively represent multi-view features, we propose efficient weight selection strategies that allow the multi-view features to capture information from the single-view features more effectively. Figure \ref{fig:pipeline} illustrates the two-stage re-identification process.

\subsection{Two-stage hierarchical person re-identification}

To query an image $q$ within a gallery set of $N$ images $G = \{g_i\}_{i=1...N}$, the distance between $q$ and each gallery image $g_i \in G$ is computed using their feature vectors. To capture essential visual information, we use a pretrained $\mathcal{F}(\cdot)$ to extract representation features. By calculating the distances between $\mathcal{F}(q)$ and each $\mathcal{F}(g_i)$, an initial ranked list $\mathcal{R}(q,G)$ = $\{ g^0_i\}_{i=1...N}$ is generated, where $d\bigl(\mathcal{F}(q),\mathcal{F}(g^0_i)\bigl) < d\bigl(\mathcal{F}(q),\mathcal{F}(g^0_{i+1})\bigl)$. Our goal is to re-rank $\mathcal{R}(q, G)$ so that more positive samples rank higher in the list, thereby improving the performance of person re-identification.


We propose a two-stage hierarchical person re-identification process as follows:

\begin{itemize}
    \item \textbf{Stage 1:} Person re-identification ranking involves ranking gallery images based on the pairwise cosine distances between the query and gallery images in the feature space of a pre-trained feature extraction model.

    \item \textbf{Stage 2:} We perform re-ranking of the top \textbf{M} candidates from Stage 1 results by computing feature distances based on multi-view features for the gallery images. Multi-view features is generated by \textit{K-nearest Weighted Fusion} proposed method.
\end{itemize}

Overall, our approach compares person images through two hierarchical stages (as depicted in Figure \ref{fig:pipeline}): single-view feature ranking stage and multi-view feature re-ranking stage.

\subsection{K-nearest Weighted Fusion for multi-view features}

\subsubsection{Generate multi-view features from K-nearest neighbors}

\begin{table*}[]
\caption{Comparison of our proposed method with various approaches on the Market-1501, MSMT17, and Occluded-DukeMTMC datasets. The results are reported in terms of Rank@1 and mAP. The best results are marked in \textbf{bold}, and the second-best results are indicated with \underline{underlining}.}
\label{table:main}
\centering
\resizebox{2.1\columnwidth}{!}{%
\begin{tabular}{lllcclcclcc}
\toprule
\multicolumn{2}{l}{}                                          &  & \multicolumn{2}{c}{\textbf{Market1501}}     & \textbf{} & \multicolumn{2}{c}{\textbf{MSMT17}}         & \textbf{} & \multicolumn{2}{c}{\textbf{Occluded-DukeMTMC}}       \\ \cmidrule{3-11} 
\multicolumn{2}{l}{}                                          &  & \textbf{Rank@1}      & \textbf{mAP}         & \textbf{} & \textbf{Rank@1}      & \textbf{mAP}         & \textbf{} & \textbf{Rank@1}      & \textbf{mAP}         \\ \midrule
\multicolumn{2}{l}{\textbf{BoT Baseline}}          &  & 94.5                 & 85.9                 &           & 74.1                 & 50.2                 &           & 48.7                 & 42.6                 \\ \midrule
\multicolumn{2}{l}{\textbf{AQE \cite{aqe}}}                   &  & 94.7                 & 91.2                 &           & 69.8                 & 55.7                 &           & 58.3                 & 55.7                 \\
\multicolumn{2}{l}{\textbf{$\alpha$-QE \cite{alphaQE}}}       &  & 95.4                 & 92.0                 &           & 70.2                 & 55.9                 &           & 57.9                 & 55.6                 \\
\multicolumn{2}{l}{\textbf{k-reciprocal \cite{k_reciprocal}}} &  & 95.4                 & 94.2                 &           & 79.8                 & \textbf{66.8}                 &           & 58.5                 & 60.3                 \\

\multicolumn{2}{l}{\textbf{ECN \cite{ecn}}} &  & 95.8                  & 93.2                 &           & -                 & -                 &           & -                 & -                 \\

\multicolumn{2}{l}{\textbf{LBR \cite{lbr}}} &  & 95.7                  & 91.3                 &           & -                 & -                 &           & -                 & -                 \\

\multicolumn{2}{l}{\textbf{GNN \cite{GNN_rerank}}}            &  & 95.8                 & \underline{94.6}        &           &               -       &      -                &           & 59.3                 & \underline{61.5}                 \\
\multicolumn{2}{l}{\textbf{GCR \cite{graph_rerank}}}            &  & \underline{96.1}                 & \textbf{94.7}        &           &               -       &      -                &           & 63.6                 & \textbf{64.3 }               \\ 
\midrule
\multirow{3}{*}{\textbf{Ours}}    & Uniform                    &  & 95.4        &     90.6             &           &    82.4             &       60.0           &           & 65.2        & 54.8                 \\

                                 & Inverse Distance Power ($p$ = 2)   &  & \multicolumn{1}{c}{\textbf{96.2}} & \multicolumn{1}{c}{87.3} &           & \multicolumn{1}{c}{\textbf{83.9}} & \multicolumn{1}{c}{56.1} &           & \multicolumn{1}{c}{\textbf{70.7}} & \multicolumn{1}{c}{52.2} \\
                                 & Exponential Decay          &  & \multicolumn{1}{c}{\underline{96.1}} & \multicolumn{1}{c}{90.8} &           & \multicolumn{1}{c}{\underline{83.2}} & \multicolumn{1}{c}{\underline{60.3}} &           & \multicolumn{1}{c}{\underline{66.9}} & \multicolumn{1}{c}{55.3} \\ \bottomrule
\end{tabular}%
}
\end{table*}

Instead of representing each image using single-view features, we propose using multi-view features to represent the images in the top \textbf{M} candidates from the initial ranked list. To generate multi-view features from single-view features, based on the methodology presented in \cite{uffm}, we propose the K-Nearest Weighted Fusion (\textit{KWF}) method, which aggregates the K-nearest neighboring features for each single-view feature. Our KWF method effectively combines neighboring features to mitigate view bias, providing a more accurate representation of individuals. Specifically, given the results from Stage 1 - $\mathcal{R}(q, G)$, we generate a list of the images in the top \textbf{M} candidates. This list of features is denoted as $\mathcal{N}(q, \textbf{M}) = \{ g^0_j\}_{j=1...M}$, where each feature of an image in $\mathcal{N}(q, \textbf{\text{M}})$ is single-view feature. With each single-view feature, $K$ neighboring features are selected and aggregated into a multi-view feature through the proposed \textit{KWF} method. This process is visualized in Figure \ref{fig:single2multi}. In our proposed \textit{KWF} method, the nearest neighboring features are selected unsupervised. This selection method may result in neighboring feature lists that include images from different classes with the single-view feature, potentially leading to multi-view features containing inaccurate information. However, theoretically, the feature extraction model $\mathcal{F}(\cdot)$ is trained such that images from the same class have high similarity, and vice versa. Thus, we expect that in the nearest neighbor feature list, the number of images from the same class as the single-view feature's image outnumber those from different classes, allowing positive features to outweigh and suppress negative ones. By recalculating the distance between $\mathcal{F}(q)$ and each multi-view feature $KWF\big(\mathcal{F}(g_j)\big)$ with $ g_j \in \mathcal{N}(q, \textbf{\text{M}})$, we obtain $\mathcal{N}^{*}(q, \textbf{\text{M}}) = \{ g^*_j\}_{j=1...M}$. By re-ranking based on multi-view features, $\mathcal{N}^{*}(q, \textbf{\text{M}})$ have more positive samples ranked higher compared to $\mathcal{N}(q, \textbf{\text{M}})$. To ensure cross-view matching settings, which is crucial in the re-identification task, the \textit{KWF} feature aggregation method excludes images in the K-nearest neighbors list with the same camera ID as the query image. It is important to note that all features used in Stage 2 are derived from Stage 1 without the need for re-extraction.

\subsubsection{Weight Selection Strategy}

When generating multi-view features, we aggregate single-view features using the \textit{KWF} method. In the study by \cite{centroids}, the aggregated feature of an image is computed by averaging the single-view features. This approach works well when the images used for aggregation belong to the same class as the single-view feature. However, in our unsupervised selection method, treating the contributions of neighboring features equally may lead to suboptimal results. Specifically, features with high similarity (mainly from the same class) and those with low similarity (possibly from different classes) contribute equally to the multi-view feature. Therefore, applying weights during the aggregation process becomes a crucial step. In this study, we thoroughly explore different weight selection strategies for aggregation, as illustrated in Figure \ref{fig:weighted}. Each weight selection strategy generates distinct multi-view features, resulting in unique aggregated representations. The visualized results demonstrate the impact of these weight selection strategies. With $f=\mathcal{F}(I)$ as the feature extracted by the model for any given image $I$, the multi-view feature $f^{(mv)}$ can be generated following:

\begin{equation}
    f^{(\text{mv})} = \sum_{k=1}^{K} w_k \boldsymbol{\cdot} f_{k}^{(\text{nn})}
\end{equation}

\begin{itemize}
    \item $f^{(mv)}$: Multi-view feature representing the image $I$
    \item $f_{k}^{(\text{nn})}$: The $k$-th neighboring feature in the list of the \textbf{K} nearest neighbors of $f$
    \item $w_k$: The aggregation weight corresponding to $f_{k}^{(\text{nn})}$

\end{itemize}

We are suggesting various methods for weight selection strategies to enhance effectiveness:

\paragraph{Uniform weighting:}This method assigns equal weights to all \textbf{K} nearest neighbor features, regardless of their distance from the single-view feature. Each neighbor has the same influence on generating the multi-view feature:

\begin{equation}
    w_k = \frac{1}{K}
\end{equation}

\paragraph{Inverse Distance Power Weighting:} In this method, weights are assigned inversely proportional to the $p$-th power of the distance between each neighboring feature $f^{(nn)}_{k}$ and the single-view feature $f$. The parameter $p$ controls the rate at which weights decrease as the distance increases. A larger $p$ emphasizes closer features more strongly, while a smaller $p$ allows for a more balanced contribution from distant features. The weights are calculated as:

\begin{equation} 
w_k = \frac{1/{d^p\bigl(f, f^{(nn)}_{k}\bigl)}}{\sum_{j=1}^{K} 1 / d^p\bigl(f, f^{(nn)}_{j}\bigl)}
\end{equation}

After extensive experimentation, we selected $p$ = 2 as our default value.

\paragraph{Exponential Decay Weighting:}This method employs an exponential function to assign weights based on distance. The weights decrease exponentially as the distance increases, ensuring that only the closest neighbors significantly influence the multi-view feature:

\begin{equation}
    w_k = \frac{e^{-d\bigl(f, f^{(nn)}_{k}\bigl)}}{\sum_{j=1}^{K}  e^{-d\bigl(f, f^{(nn)}_{j}\bigl)}}
\end{equation}

\begin{table*}[]
\caption{Comparison of the computation time and memory usage in re-ranking methods}
\label{tab:computing}
\centering
\resizebox{2.1\columnwidth}{!}{%
\begin{tabular}{lcccc}
\toprule
\multirow{2}{*}{\textbf{Method}} & \multicolumn{2}{c}{\textbf{Market1501}}                & \multicolumn{2}{c}{\textbf{Occluded-DukeMTMC}}         \\ \cmidrule{2-5} 
                                 & \textbf{Evaluate time ($\downarrow$)} & \textbf{GPU Memory ($\downarrow$)} & \textbf{Evaluate time ($\downarrow$)} & \textbf{GPU Memory ($\downarrow$)} \\ \midrule
\textbf{AQE \cite{aqe}}                     & \textcolor{white}{00}\textbf{7.8s}                       & 10.55GB                  & \textcolor{white}{00}8.8s                      & 12.53GB                   \\
\textbf{$\alpha$-QE \cite{alphaQE}}                 & \textcolor{white}{00}7.9s                        & 10.55GB                  & \textcolor{white}{00}8.8s                       & 12.53GB                   \\
\textbf{k-reciprocal \cite{k_reciprocal}}                  & 146.0s                     &                 \textcolor{white}{0}5.63GB          & 150.8s                     &    \textcolor{white}{0}5.62GB                       \\
\textbf{GNN \cite{GNN_rerank}}                     & \textcolor{white}{00}8.6s                       & \textcolor{white}{0}4.75GB                   & 10.97s                      & \textcolor{white}{0}4.96GB                    \\
\textbf{GCR \cite{graph_rerank}}                     &           \textcolor{white}{0}34.5s            &   20.97GB                 &               \textcolor{white}{0}35.7s        & 23.24GB                   \\ \midrule
\textbf{Our}                     & \textcolor{white}{00}8.5s                       & \textcolor{white}{0}\textbf{1.16GB}                   & \textcolor{white}{00}\textbf{6.1s}                       & \textcolor{white}{0}\textbf{1.11GB}                   \\ \bottomrule
\end{tabular}%
}
\end{table*}

\begin{table*}[]
\centering
\caption{Performance of our method on Market-1501 and Occluded-DukeMTMC across different top \textbf{M} candidates.}
\label{table:M}
\begin{tabular}{llllllccccllllcccc}
\toprule
 & \multirow{2}{*}{\textbf{Top-M}} &  &  &  &  & \multicolumn{4}{c}{\textbf{Market-1501}}              &  &  &  &  & \multicolumn{4}{c}{\textbf{Occluded-DukeMTMC}}        \\ \cmidrule{6-18} 
 &                                 &  &  &  &  & \textbf{Rank-1} & \textbf{mAP} & \textbf{Time(ms)} &  &  &  &  &  & \textbf{Rank-1} & \textbf{mAP} & \textbf{Time(ms)} &  \\ \midrule
 & \textbf{20}                     &  &  &  &  & 96.0            & 88.3         & 1.41              &  &  &  &  &  & 58.9            & 46.7         & 1.58              &  \\
 & \textbf{40}                     &  &  &  &  & 96.1            & 89.8         & 1.47              &  &  &  &  &  & 62.6            & 50.6         & 1.70              &  \\
 & \textbf{60}                     &  &  &  &  & 96.1            & 90.3         & 1.59              &  &  &  &  &  & 64.7            & 52.8         & 1.73              &  \\
 & \textbf{80}                     &  &  &  &  & 96.1            & 90.6         & 1.70              &  &  &  &  &  & 66.1            & 54.2         & 1.85              &  \\
 & \textbf{100}                    &  &  &  &  & 96.1            & 90.8         & 1.76              &  &  &  &  &  & 66.9            & 55.3         & 1.93              &  \\
 & \textbf{120}                    &  &  &  &  & 96.1            & 90.9         & 1.88              &  &  &  &  &  & 67.6            & 56.1         & 2.03              &  \\
 & \textbf{140}                    &  &  &  &  & 96.1            & 90.9         & 1.94              &  &  &  &  &  & 68.1            & 56.8         & 2.17              &  \\
 & \textbf{160}                    &  &  &  &  & 96.1            & 91.0         & 2.03              &  &  &  &  &  & 68.6            & 57.4         & 2.25              &  \\ \bottomrule
\end{tabular}%
\end{table*}

\section{\uppercase{Results}}

\subsection{Datasets and Settings}
Our main results are trained and evaluated on three-person re-identification datasets:
\begin{itemize}
    \item \textbf{Market-1501} \cite{market1501}\textbf{:} The Market-1501 dataset, gathered at Tsinghua University, features 32,668 images of 1,501 individuals captured by six cameras positioned outside a supermarket. The dataset is split into 12,936 training images from 751 individuals and 19,732 testing images from 750 individuals, with 3,368 of the test images designated as the query set.
    \item \textbf{MSMT17} \cite{msmt17}\textbf{:} MSMT17 is an extensive dataset for multi-scene and multi-time person re-identification. It comprises 180 hours of video from 12 outdoor and 3 indoor cameras across 12 different time slots. The dataset includes 4,101 annotated identities and 126,441 bounding boxes, providing a robust foundation for re-identification studies.
    \item \textbf{Occluded-DukeMTMC} \cite{occduke}\textbf{:} This dataset contains 15,618 training images, 17,661 gallery images, and 2,210 occluded query images. Our experiments on the Occluded-DukeMTMC dataset demonstrate the efficacy of our method in handling occluded person re-identification. Importantly, our approach does not require manual cropping during preprocessing.
\end{itemize}

\begin{figure*}[ht]
    \centering

    \begin{subfigure}[b]{1.0\textwidth}
        \centering
        \includegraphics[width=\textwidth]{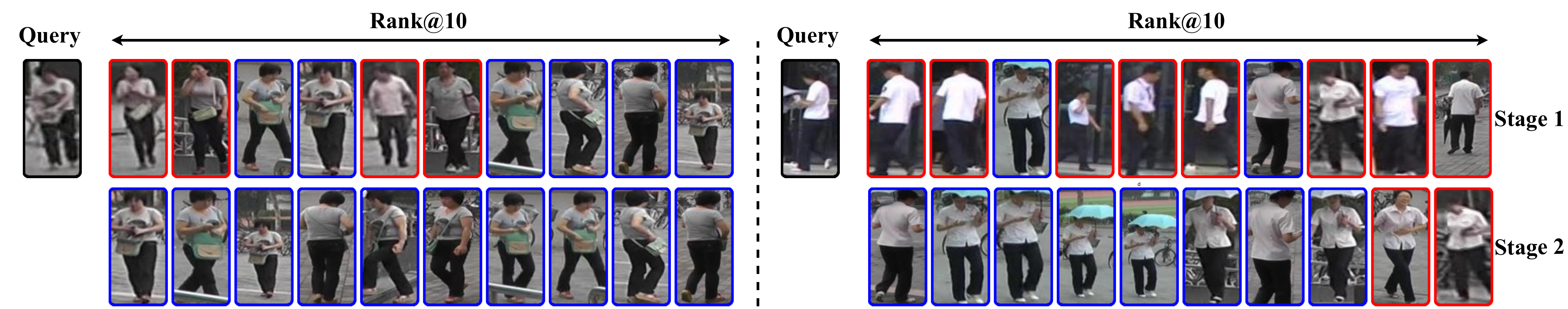} 
        \caption{Market1501}
        \label{fig:main_sub2}
    \end{subfigure}
    \begin{subfigure}[b]{1.0\textwidth}
        \centering
        \includegraphics[width=\textwidth]{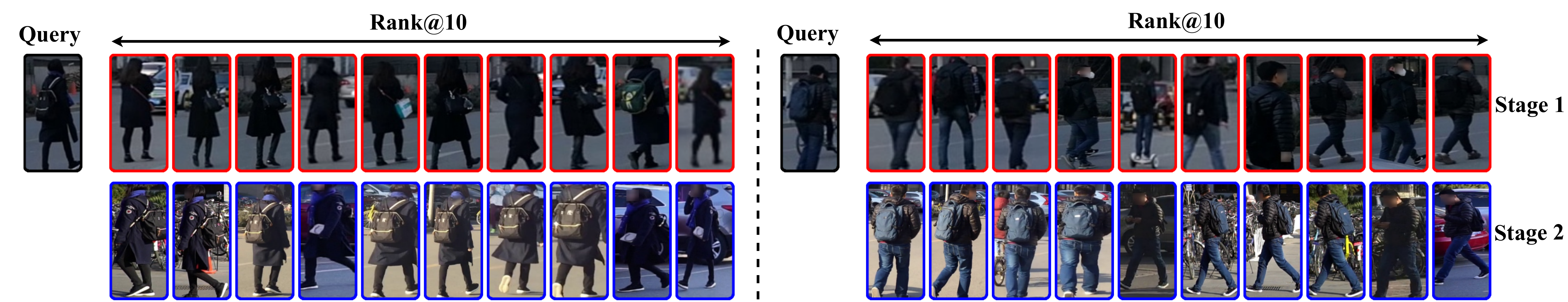} 
        \caption{MSMT17}
        \label{fig:main_sub3}
    \end{subfigure}
    \begin{subfigure}[b]{1.0\textwidth}
        \centering
        \includegraphics[width=\textwidth]{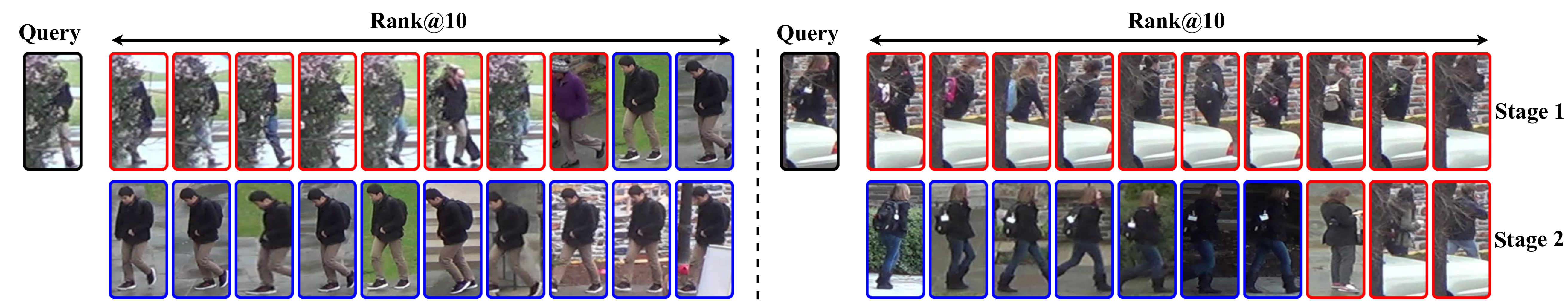} 
        \caption{Occluded-DukeMTMC}
        \label{fig:main_sub4}
    \end{subfigure}

    \caption{The results from the experiments demonstrate the effectiveness of the proposed reranking method. In each sub-figure (Market1501, MSMT17, and Occluded-DukeMTMC), the query images are on the left, followed by columns showing the top 10 most retrieved results. The results from Stage 1 are on the top row of each dataset, while the reranking results from Stage 2 are on the bottom row. \textcolor{blue}{Blue} and \textcolor{red}{red} boxes indicate true positives and false positives, respectively.
}
    \label{fig:visualize}
    
\end{figure*}

\begin{figure*}[ht]
    \centering
    \begin{subfigure}[b]{0.46\textwidth}
        \centering
        \includegraphics[width=\textwidth]{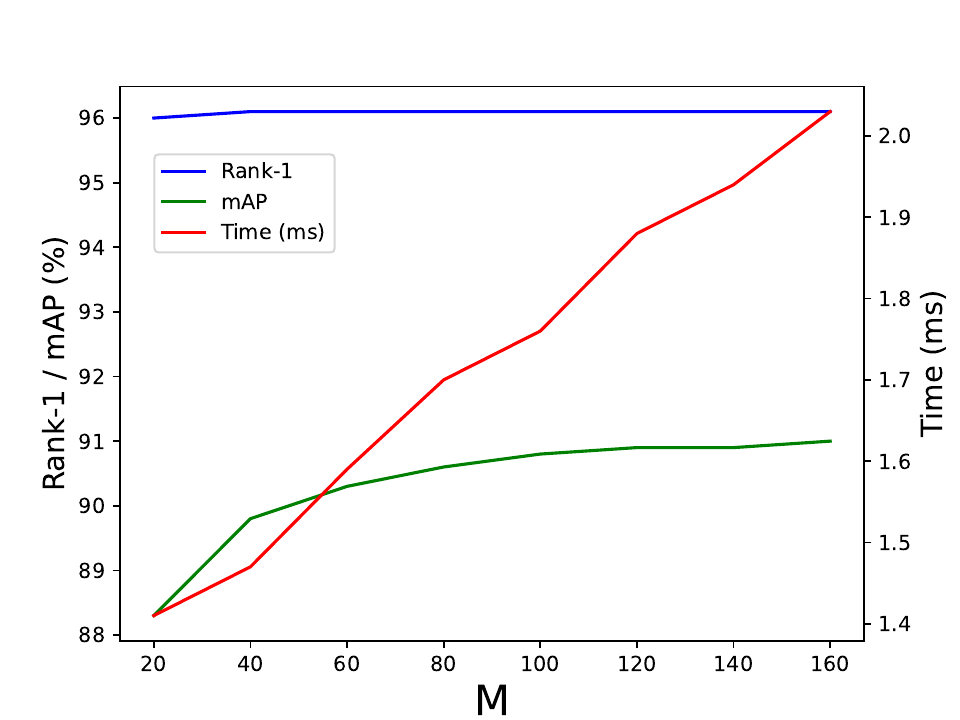} 
        \caption{Market-1501}
        \label{fig:sub1}
    \end{subfigure}
    \begin{subfigure}[b]{0.46\textwidth}
        \centering
        \includegraphics[width=\textwidth]{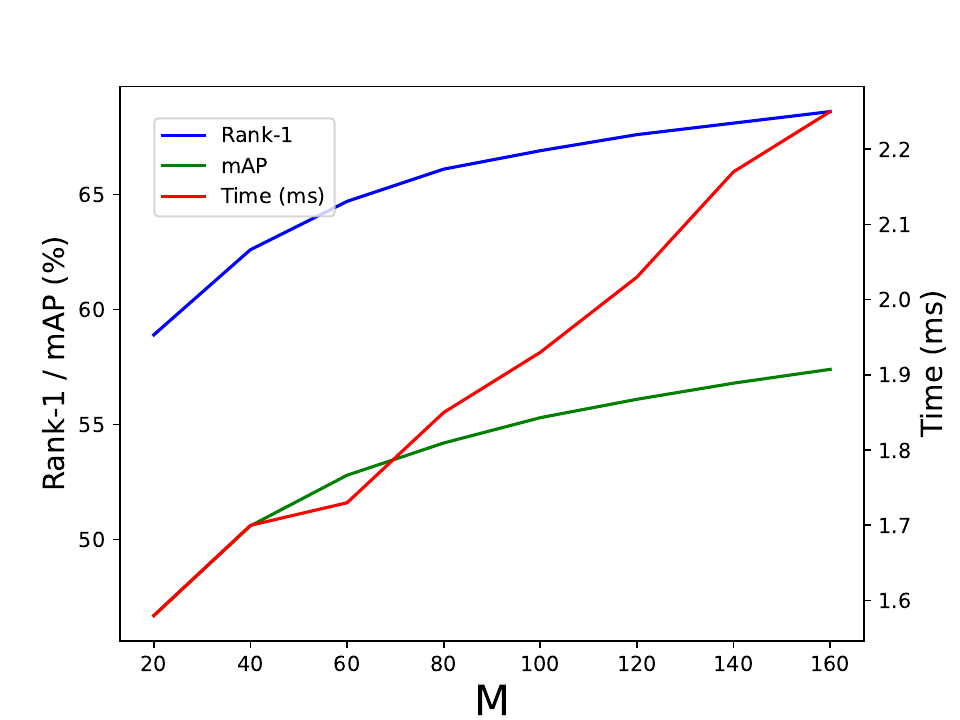} 
        \caption{Occluded-DukeMTMC}
        \label{fig:sub2}
    \end{subfigure}
    \caption{Results when select different top \textbf{M} candidates.}
    \label{fig:M}
\end{figure*}

\begin{figure*}[ht]
    \centering
    \begin{subfigure}[b]{0.46\textwidth}
        \centering
        \includegraphics[width=\textwidth]{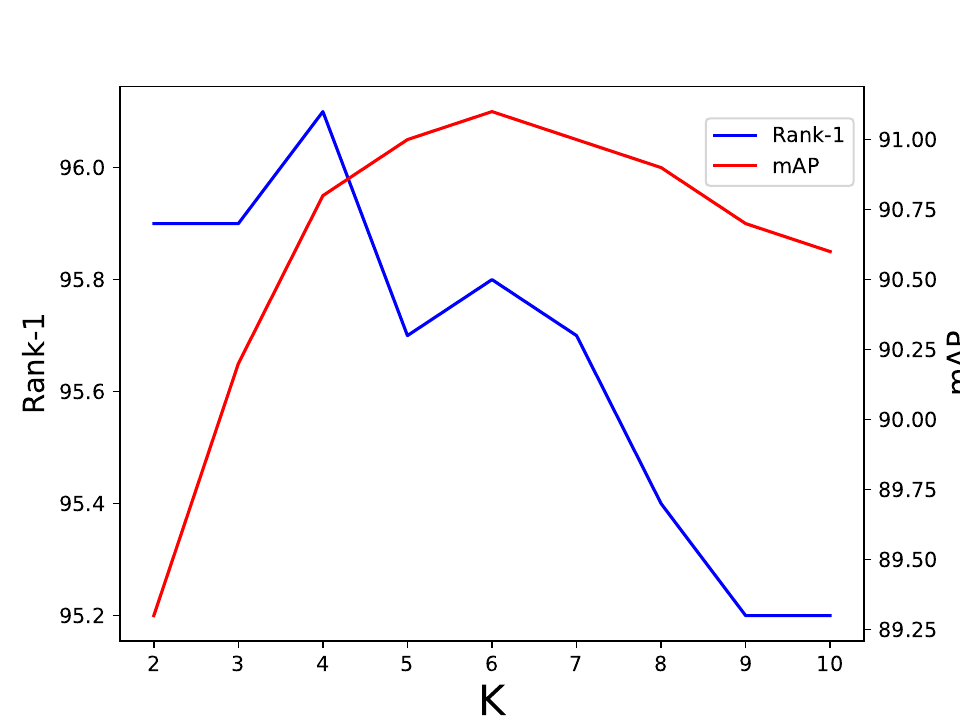} 
        \caption{Market-1501}
    \end{subfigure}
    \begin{subfigure}[b]{0.46\textwidth}
        \centering
        \includegraphics[width=\textwidth]{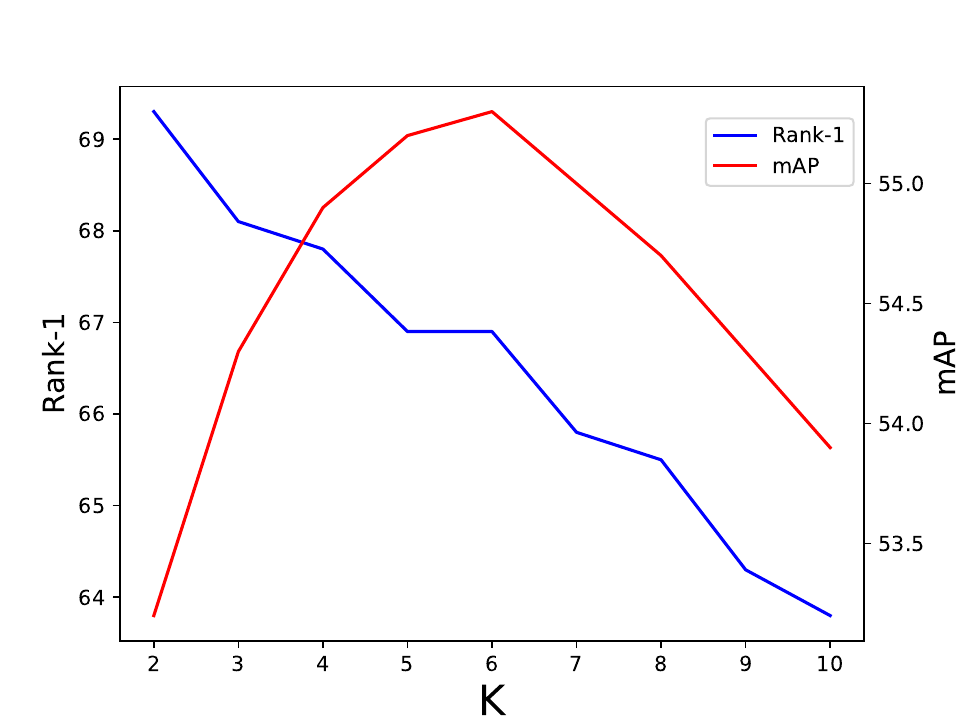} 
        \caption{Occluded-DukeMTMC}
    \end{subfigure}
    \caption{Results when changing \textbf{K} for finding nearest neighbors.}
    \label{fig:K}
\end{figure*}

We utilize the BoT \cite{bot} with a ResNet-50 backbone to extract features while evaluating our proposed. We use two evaluation metrics to assess the performance of person Re-ID methods across all datasets. The first metric is the Cumulative Matching Characteristic (CMC). Viewing re-ID as a ranking problem, we report the cumulative matching accuracy at Rank@1. The second metric is the mean Average Precision (mAP). We evaluate the proposed re-ranking method on the top 100 candidates in the initial result list (\textbf{M} = 100). For \textbf{K}, which is the number of nearest neighbors, we select \textbf{K} = 6 for the MSMT17 and Occluded-DukeMTMC datasets, while for Market1501, we choose \textbf{K} = 4. The selection of these hyperparameters \textbf{M} and \textbf{K} is further analyzed in Section \ref{ab}. All experiments used the PyTorch framework on an GeForce RTX 4090 GPU with 32GB of RAM and an Intel(R) Core(TM) i9-10900X processor.
\subsection{Main results}
\subsubsection{Quantitative results}

As shown in Table \ref{table:main}, our method \textbf{consistently outperforms
the traditional Stage 1 alone} in both Rank@1 and mAP metrics, regardless of the weight selection strategies used. Among the three strategies, \textit{Inverse Distance Power Weighting ($p$ = 2)} demonstrates the best Rank@1 performance, while \textit{Exponential Decay Weighting} shows uniform improvement in both Rank@1 and mAP. Specifically, with \textit{Inverse Distance Power Weighting}, our method achieves improvements of \textbf{1.7\%}, \textbf{9.8\%}, and \textbf{22.0\%} in Rank@1 on the Market-1501, MSMT17, and Occluded-DukeMTMC datasets, respectively. Moreover, when comparing our method to other post-processing methods, it delivers higher Rank@1 accuracy, particularly on the MSMT17 and Occluded-DukeMTMC datasets. Our approach focuses on re-ranking the top \textbf{M} candidates based on multi-view features, making it particularly effective when evaluated on the Rank@1 metric. Our method improves person re-identification, especially in datasets with occlusions and complex variations, such as MSMT17 and Occluded-DukeMTMC. The proposed weight selection strategies enhance feature representation, leading to better identity discrimination.

\begin{table}[]
\centering
\caption{Performance of our proposed on Market-1501 and Occluded-DukeMTMC across different \textbf{K} values.}
\label{table:K}
\begin{tabular}{llcccccc}
\toprule
\multirow{2}{*}{\textbf{K}} &  & \multicolumn{2}{c}{\textbf{Market-1501}} & \multicolumn{2}{c}{\textbf{Occluded-DukeMTMC}} \\ \cmidrule{3-6} 
                                 &  & \textbf{Rank-1} & \textbf{mAP} & \textbf{Rank-1} & \textbf{mAP} \\ \hline
\textbf{2}                      &  & 95.9            & 89.3         & 69.3            & 53.2        \\
\textbf{3}                      &  & 95.9            & 90.2         & 68.1            & 54.3        \\
\textbf{4}                      &  & 96.1            & 90.8         & 67.8            & 54.9        \\
\textbf{5}                      &  & 95.7            & 91.0         & 66.9            & 55.2        \\
\textbf{6}                      &  & 95.8            & 91.1         & 66.9            & 55.3        \\
\textbf{7}                      &  & 95.7            & 91.0         & 65.8            & 55.0        \\
\textbf{8}                      &  & 95.4            & 90.9         & 65.5            & 54.7        \\
\textbf{9}                      &  & 95.2            & 90.7         & 64.3            & 54.3        \\
\textbf{10}                     &  & 95.2            & 90.6         & 63.8            & 53.9        \\ \hline
\end{tabular}
\end{table}

To compare the computational cost of our method with other approaches, we evaluate memory usage and evaluation time (measuring the time five times and taking the average result). All methods are implemented on a GPU (except for the k-reciprocal \cite{k_reciprocal}, executed on the GPU in Stage 1 and the CPU in Stage 2). We do not account for the image feature extraction time; instead, we only measure each method's time to process queries and perform re-ranking. The results in Table \ref{tab:computing} demonstrate that our method requires significantly less memory than other methods, utilizing only about 1GB of GPU memory. Our approach involves only the time needed to generate multi-view features without requiring additional memory to store new features or auxiliary components. Additionally, with competitive processing time, our method shows strong applicability for large-scale retrieval systems.

\subsubsection{Qualitative results}

Figure \ref{fig:visualize} visualizes some results after using our re-ranking method. Stage 1 shows the initial results based on direct feature distance computation, while Stage 2 presents the re-ranking results using multi-view features. The outcomes demonstrate the robustness of our method in re-identifying images with changes in viewpoint and background across the Market1501 and MSMT17 datasets. Even more impressively, on the Occluded-DukeMTMC dataset, our re-identification approach accurately handles cases where the persons are occluded. These results show that our method effectively aggregates information, significantly enhancing retrieval performance during re-ranking.

\begin{figure*}[ht]
    \centering
    \begin{subfigure}[b]{0.46\textwidth}
        \centering
        \includegraphics[width=\textwidth]{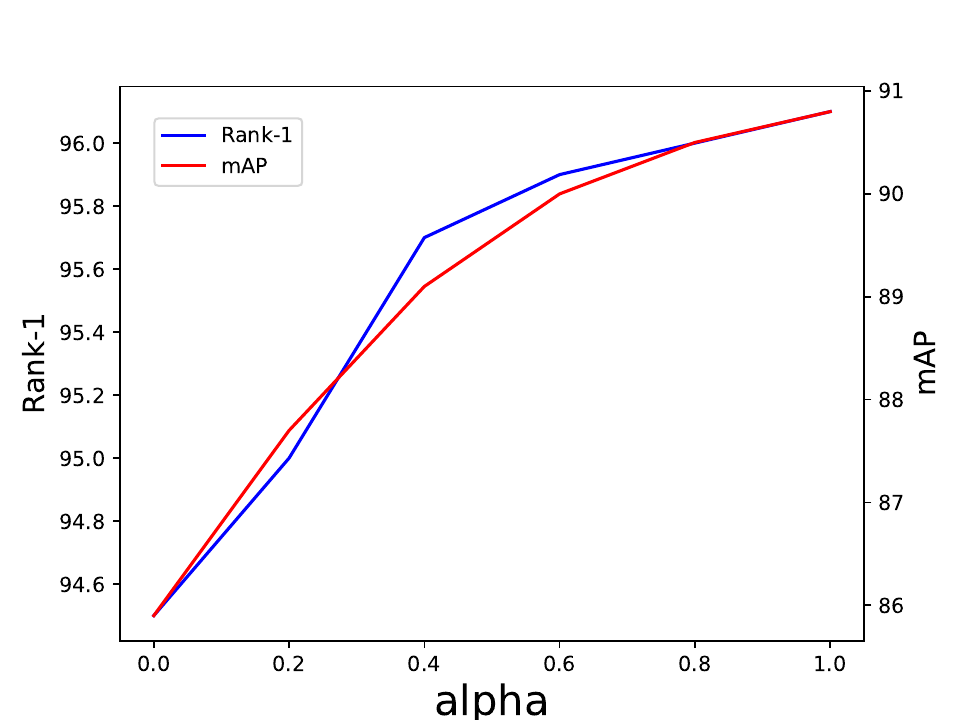} 
        \caption{Market-1501}
    \end{subfigure}
    \begin{subfigure}[b]{0.46\textwidth}
        \centering
        \includegraphics[width=\textwidth]{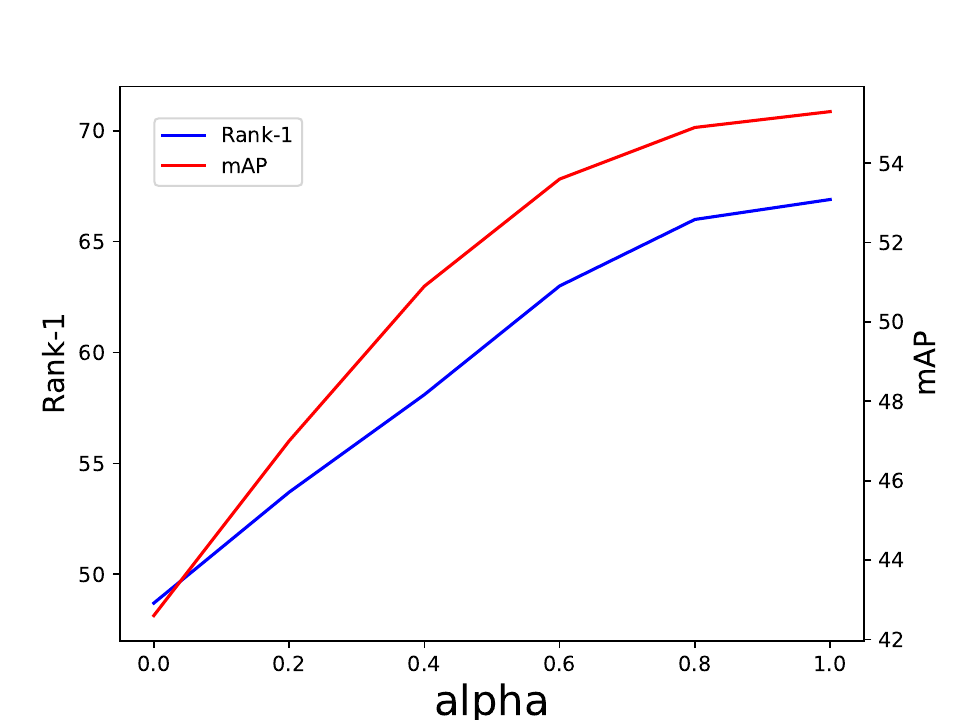} 
        \caption{Occluded-DukeMTMC}
    \end{subfigure}
    \caption{Results when sweeping across $\alpha$ for linearly combining multi-view and single-view features}
    \label{fig:alpha}
\end{figure*}

\subsection{Ablation Studies and Analysis} \label{ab}

In this section, we perform extensive ablation studies on the Market1501 and Occluded-DukeMTMC datasets to examine the impact and sensitivity of various hyperparameters.

\subsubsection{Effect of different \textbf{K}}

To analyze the impact of the number of nearest neighbors \textbf{K} we conducted experiments on the Market-1501 and Occluded-DukeMTMC datasets. Table \ref{table:K} and Figure \ref{fig:K} demonstrate the trade-off between Rank@1 and mAP as \textbf{K} increases from 2 to 10. Specifically, Rank@1 tends to decrease as \textbf{K} increases, while mAP rises to \textbf{K} = 6 , after which it starts to decline. For the Occluded-DukeMTMC dataset, we chose \textbf{K} = 6 as it provides the best balance between Rank-1 and mAP. For the Market-1501 dataset, we selected \textbf{K} = 4 to maintain Rank-1 performance, as this metric decreases rapidly with increasing \textbf{K}.

\begin{table}[]
\centering
\caption{Performance of our proposed on Market-1501 and Occluded DukeMTMC across different $\alpha$ values.}
\label{table:alpha}
\begin{tabular}{llcccccc}
\toprule
\multirow{2}{*}{\bm{$\alpha$}} &  & \multicolumn{2}{c}{\textbf{Market-1501}} & \multicolumn{2}{c}{\textbf{Occluded-DukeMTMC}} \\ \cmidrule{3-6} 
                                   &  & \textbf{Rank-1} & \textbf{mAP} & \textbf{Rank-1} & \textbf{mAP} \\ \hline
\textbf{0.0}                       &  & 94.5            & 85.9         & 48.7            & 42.6        \\
\textbf{0.2}                       &  & 95.0            & 87.7         & 53.7            & 47.0        \\
\textbf{0.4}                       &  & 95.7            & 89.1         & 58.1            & 50.9        \\
\textbf{0.6}                       &  & 95.9            & 90.0         & 63.0            & 53.6        \\
\textbf{0.8}                       &  & 96.0            & 90.5         & 66.0            & 54.9        \\
\textbf{1.0}                       &  & 96.1            & 90.8         & 66.9            & 55.3        \\ \hline
\end{tabular}
\end{table}

\subsubsection{Effect of different top \textbf{M} candidates}

The selection of the top \textbf{M} candidates, as observed in Table \ref{table:M} and Figure \ref{fig:M}, illustrates the trade-off between accuracy and query time. The query time refers to the duration required to perform a single query, measured five times, with the average value reported. Choosing a higher \textbf{M} value means re-ranking more candidates using multi-view features, which improves accuracy but also requires more time for feature aggregation. Therefore, we choose \textbf{M} = 100 in this study to balance accuracy and query time.

\subsubsection{Combining multi-view and single-view features.}
In this section, we explore combining the generated multi-view feature and the original single-view feature by linearly combining $f$ and $f^{(mv)}$ using $\alpha \in \{$0.0, 0.2, 0.4,..., 1.0$\}$:

\begin{equation}
    f^{*} = (1-\alpha) \times f + \alpha \times f^{(mv)}
\end{equation}

We found that varying $\alpha$ impacts both Rank@1 and mAP results across the Market-1501 and Occluded-DukeMTMC datasets. The results in Figure \ref{fig:alpha} and Table \ref{table:alpha} show that the larger the value of $\alpha$, the higher the performance. This indicates that the more significant the contribution of multi-view features, the better the re-ranking performance. Therefore in all experiments we choose $\alpha$ = 1 and ignore the contribution of $f$.

\subsubsection{Effectiveness in different baselines}

In addition to the results on the baseline BoT Resnet-50, we also experimented with our method on different backbones. Table \ref{tab:backbone} shows the results on various baselines, including BoT Resnet101-ibn \cite{bot}, CLIP-ReID Resnet-50, and ViT-B/16 \cite{clip-reid}. The results demonstrate that our method can be easily applied to other pre-trained models, leading to significant improvements in both Rank@1 and mAP.

\begin{table*}[]
\caption{The results of our method applied to different baselines.}

\label{tab:backbone}
\begin{tabular}{lcccccc}
\toprule
\multicolumn{1}{c}{\multirow{2}{*}{\textbf{Method}}} & \multicolumn{2}{c}{\textbf{Market1501}}                                                                     & \multicolumn{2}{c}{\textbf{Occluded-DukeMTMC}}                                                               & \multicolumn{2}{c}{\textbf{MSMT17}}                                                                         \\ \cmidrule{2-7} 
\multicolumn{1}{c}{}                                 & \textbf{Rank@1}                                      & \textbf{mAP}                                         & \textbf{Rank@1}                                      & \textbf{mAP}                                          & \textbf{Rank@1}                                      & \textbf{mAP}                                         \\ \midrule
CLIP-ReID (CNN)                                      & 94.7                                                 & 88.1                                                 & 54.2                                                 & 47.4                                                  & -                                                    & -                                                    \\
+ \textbf{Ours}                                      & 95.7 {\color[HTML]{3531FF}($\uparrow$ \textbf{1.0})} & 91.8 {\color[HTML]{3531FF}($\uparrow$ \textbf{3.7})} & 68.1 {\color[HTML]{3531FF}($\uparrow$ \textbf{13.9})} & 59.5 {\color[HTML]{3531FF}($\uparrow$ \textbf{12.1})} & -                                                    & -                                                    \\
\hdashline CLIP-ReID (ViT)                           & 93.3                                                 & 86.4                                                 & 60.8                                                 & 53.5                                                  & -                                                    & -                                                    \\
+ \textbf{Ours}                                      & 93.7 {\color[HTML]{3531FF}($\uparrow$ \textbf{0.4})}          & 87.8 {\color[HTML]{3531FF}($\uparrow$ \textbf{1.4})} & 64.8 {\color[HTML]{3531FF}($\uparrow$ \textbf{4.0})} & 60.1 {\color[HTML]{3531FF}($\uparrow$ \textbf{6.6})}  & -                                                    & -                                                    \\
\hdashline BoT (R101-ibn)                            & 95.4                                                 & 88.9                                                 & -                                                    & -                                                     & 81.0                                                 & 59.4                                                 \\
\textbf{+ Ours}                                      & 96.1 {\color[HTML]{3531FF}($\uparrow$ \textbf{1.7})} & 92.2 {\color[HTML]{3531FF}($\uparrow$ \textbf{3.3})} & -                                                    & -                                                     & 86.3 {\color[HTML]{3531FF}($\uparrow$ \textbf{5.3})} & 67.5 {\color[HTML]{3531FF}($\uparrow$ \textbf{8.1})} \\ \bottomrule
\end{tabular}%
\end{table*}

\begin{table*}[]
\centering
\caption{Experimental results evaluating our method with different indexing techniques, highlighting the trade-offs between query time and re-identification accuracy.}
\label{tab:index}
\begin{tabular}{lcccclcccc}
\toprule
\multicolumn{1}{c}{} & \multicolumn{4}{c}{\textbf{Market-1501}} &  & \multicolumn{4}{c}{\textbf{Occluded-DukeMTMC}} \\ \cmidrule{2-10} 
\multicolumn{1}{c}{} & \multicolumn{2}{c}{\textbf{Time (s)}} &  &  &  & \multicolumn{2}{c}{\textbf{Time (s)}} &  &  \\ \cmidrule{2-3} \cmidrule{7-8}
\multicolumn{1}{c}{\multirow{-3}{*}{\textbf{Type}}} & \textbf{CPU} & \textbf{GPU} & \multirow{-2}{*}{\textbf{Rank@1}} & \multirow{-2}{*}{\textbf{mAP}} &  & \textbf{CPU} & \textbf{GPU} & \multirow{-2}{*}{\textbf{Rank@1}} & \multirow{-2}{*}{\textbf{mAP}} \\ \midrule
\textbf{Normal} & 597.21 & 8.46 & 96.1 & 90.8 &  & 423.26 & 6.02 & 69.1 & 55.3 \\ \midrule
{\color[HTML]{242424} \textbf{IndexIVFPQ}} & \begin{tabular}[c]{@{}c@{}}357.84\\ {\color[HTML]{3531FF} $\downarrow$ 40.08\%}\end{tabular} & \begin{tabular}[c]{@{}c@{}}7.76\\ {\color[HTML]{3531FF} $\downarrow$ 8.27\%}\end{tabular} & \begin{tabular}[c]{@{}c@{}}95.1\\ {\color[HTML]{FE0000} $\downarrow$ 1.0}\end{tabular}  & \begin{tabular}[c]{@{}c@{}}88.4\\ {\color[HTML]{FE0000} $\downarrow$ 2.4}\end{tabular} & & \begin{tabular}[c]{@{}c@{}}237.31\\ {\color[HTML]{3531FF} $\downarrow$ 43.93\%}\end{tabular} & \begin{tabular}[c]{@{}c@{}}5.11\\ {\color[HTML]{3531FF} $\downarrow$ 15.12\%}\end{tabular} & \begin{tabular}[c]{@{}c@{}}62.3\\ {\color[HTML]{FE0000} $\downarrow$ 6.8}\end{tabular} & \begin{tabular}[c]{@{}c@{}}51.7\\ {\color[HTML]{FE0000} $\downarrow$ 3.6}\end{tabular} \\ \hdashline
{\color[HTML]{242424} \textbf{IndexIVFFlat}} & \begin{tabular}[c]{@{}c@{}}354.63\\ {\color[HTML]{3531FF} $\downarrow$ 40.62\%}\end{tabular} & \begin{tabular}[c]{@{}c@{}}7.19\\ {\color[HTML]{3531FF} $\downarrow$ 15.01\%}\end{tabular} & \begin{tabular}[c]{@{}c@{}}95.5\\ {\color[HTML]{FE0000} $\downarrow$ 0.6}\end{tabular} & \begin{tabular}[c]{@{}c@{}}89.0\\ {\color[HTML]{FE0000} $\downarrow$ 1.8}\end{tabular} &  & \begin{tabular}[c]{@{}c@{}}232.42\\ {\color[HTML]{3531FF} $\downarrow$ 45.09\%}\end{tabular} & \begin{tabular}[c]{@{}c@{}}4.74\\ {\color[HTML]{3531FF} $\downarrow$ 21.26\%}\end{tabular} & \begin{tabular}[c]{@{}c@{}}61.4\\ {\color[HTML]{FE0000} $\downarrow$ 7.7}\end{tabular} & \begin{tabular}[c]{@{}c@{}}51.4\\ {\color[HTML]{FE0000} $\downarrow$ 3.9}\end{tabular} \\ \hdashline
\textbf{IndexLSH} & \begin{tabular}[c]{@{}c@{}}515.09\\ {\color[HTML]{3531FF} $\downarrow$ 13.75\%}\end{tabular} & - & \begin{tabular}[c]{@{}c@{}}96.2\\ {\color[HTML]{3531FF} $\uparrow$ 0.1}\end{tabular} & \begin{tabular}[c]{@{}c@{}}88.0\\ {\color[HTML]{FE0000} $\downarrow$ 2.8}\end{tabular} &  & \begin{tabular}[c]{@{}c@{}}358.89\\ {\color[HTML]{3531FF} $\downarrow$ 15.21\%}\end{tabular} & - & \begin{tabular}[c]{@{}c@{}}70.7\\ {\color[HTML]{3531FF} $\uparrow$ 1.6}\end{tabular} & \begin{tabular}[c]{@{}c@{}}53.2\\ {\color[HTML]{FE0000} $\downarrow$ 2.1}\end{tabular} \\ \bottomrule
\end{tabular}%
\end{table*}

\subsubsection{Time cost comparison with different feature index methods}

In real-world applications such as surveillance and security, indexing is crucial in managing large-scale galleries. Most available indexing structures involve a trade-off between query time and accuracy. Therefore, in this study, we evaluate our method using various indexing types \cite{faiss_gpu,faiss}. Table \ref{tab:index} presents our experimental results. Across both datasets, using IndexIVFPQ and IndexIVFFlat demonstrates a similar trade-off: a slight decrease in re-identification accuracy but a significant reduction in evaluation time on the CPU. On the other hand, when using IndexLSH, the time improvement is not substantial; however, there is a minor increase in Rank@1 accuracy compared to the standard approach. These experiments provide deeper insights into the practical application of our method in real-world query systems.

\section{\uppercase{Discussion and Conclusion}}
\subsection{Limitations}
In this study, we primarily focus on exploring the potential of a multi-view feature-based approach for the re-ranking stage without optimizing the re-ranking method, resulting in less competitive mAP scores. Additionally, since our methods select features in an unsupervised manner, the performance depends on the pre-trained models. Therefore, the performance of our method relies on the performance of the pre-trained person re-identification models. Finally, we have yet to investigate the effectiveness of multi-view features in other retrieval tasks, which could be an exciting direction for future research.

\subsection{Conclusion}

In this study, we proposed a two-stage hierarchical person re-identification approach combining single-view and multi-view features. Introducing the K-nearest Weighted Fusion (\textit{KWF}) method addressed the challenges posed by view bias and significantly improved re-ranking performance without requiring additional fine-tuning or annotations. Experimental results on Market-1501, MSMT17, and Occluded-DukeMTMC datasets demonstrate that our method outperforms existing re-ranking techniques in Rank-1 accuracy while maintaining computational efficiency. Our approach improved substantially on challenging datasets with occlusions, highlighting its robustness and practical applicability. This work advances the re-ranking in person re-identification and opens avenues for future research in adaptive feature aggregation and the application of multi-view representations to other domains. By optimizing feature representation, we aim to contribute to developing more accurate and efficient retrieval systems for real-world applications.

\section{\uppercase{Acknowledgments}}

This research is funded by University of Information Technology-Vietnam National University of Ho Chi Minh city under grant number D1-2024-70.



\bibliographystyle{apalike}
{\small
\bibliography{ref}}




\end{document}